# Implementation of Deep Convolutional Neural Network in Multi-class Categorical Image Classification

Pushparaja Murugan

*School of Mechanical and Aerospace Engineering,*

*Nanyang Technological University, Singapore 639815.*

(pushpara001@e.ntu.edu.sg )

## Abstract

Convolutional Neural Networks has been implemented in many complex machine learning takes such as image classification, object identification, autonomous vehicle and robotic vision tasks. However, ConvNet architecture's efficiency and accuracy depend on a large number of factors. Also, the complex architecture requires a significant amount of data to train and involves with a large number of hyperparameters that increases the computational expenses and difficulties. Hence, it is necessary to address the limitations and techniques to overcome the barriers to ensure that the architecture performs well in complex visual tasks. This article is intended to develop an efficient ConvNet architecture for multi-class image categorical classification application. In the development of the architecture, large pool of grey scale images are taken as input information images and split into training and test datasets. The numerously available technique is implemented to reduce the overfitting and poor generalization of the network. The hyperparameters of determined by Bayesian Optimization with Gaussian Process prior algorithm. ReLu non-linear activation function is implemented after the convolutional layers. Max pooling operation is carried out to downsampling the data points in pooling layers. Cross-entropy loss function is used to measure the performance of the architecture where the softmax is used in the classification layer. Mini-batch gradient descent with Adam optimizer algorithm is used for backpropagation. Developed architecture is validated with confusion matrix and classification report.

# 1 Introduction

Convolutional Neural Network (CNN) is biologically inspired MLP networks and is developed based on mathematical representation to solve many image visual imagery application, object classification and speech recognition. The study of neural network has started in early 1950. Different types of neural networks are successfully developed such as Elman, Hopfield and Jordan networks for approximating complex functions and recognizing patterns in the late 1970s. [1] [2] [3]. Hubel and Wiesel studies on cat visual cortex system have inspired the researcher to develop an artificial pattern recognition system. In their study, they proposed that the visual cortex is sensitive to small sub-regions of the visual field, known as a receptive field. The simple cells in cortex system respond to edge like patterns, and larger receptive fields are invariant to the position of the natural images [4]. Based on this local sensitive and orientation-selective neurons, ConvNets are introduced by Fukushima to solve pattern recognition problem [5] and later developed for identifying the digital characters by LeCun [6]. From this respective local field, neurons are extracted the elementary features and combine them in subsequent layers to get the complex features of the original images. The development of AlexNet in 2012 by Krizhevsky is achieved high-level performance and outperformed all other networks in ImageNet-2012 competition where they trained the 15 million images to classify 2200 classes [7]. After the successful implementation of ConvNet, the architecture is used in many in object classification, natural language processing and pattern recognition, cancer cell classification, medical image processing application, star cluster classification, self-driving cars and number plate recognition.

Unlike machine learning, CovnNets can be fed with raw image pixel values rather than feature vectors as input [8]. The fundamental design principle of CovnNets is developing an architecture and learning algorithm in such way that it reduces the number of the parameter without compromising the computational power of learning algorithm [9]. There are four main ideas in the development of ConvNet architecture: Local connection, sharing the weights, pooling the data and implementation of multilayers [8]. The typical architecture of ConvNets consists of several layers. There are convolutional layers, pooling layers and fully connected dense layers [10]. The convolution processes act as suitable feature detectors that demonstrate the ability to deal with a large amount of low-level information. A complete convolution layer has different feature detectors so that multiple features can be extracted from the same image. A single feature detector is smaller in size as compares with the input images is slid over the images for the convolution operation. Hence, all of the units in that feature detector share the same weight and bias. That will help to detect same features in all of the points in the image. That gives the properties of invariance to transformation and shift of the images [11]. Local connections between the pixels are used many times in architecture. With local respective field, neurons can extract the simple features such as the orientation of edges and corners and end points. So that higher degree of complex features is detected in hidden layers when its combined in hidden layers. These functions of sparse connectivity between subsequent layers, parameter sharing of weights between the adjacent pixels and equivalent representation enable CNN to use efficiently in image reorganization and image classification problems [10] [12].

However, learning of deep network architecture demands a significant amount of data and learning from the input data is a highly computational demanding task. The more substantial number of connections between the neurons and the parameters are tuned on iterative basis over the cost function by gradient descent optimization or its variants. Also, sometimes developed architecture tends to suffer from overfitting the test data. In this situation, regularization and optimization techniques could be used to overcome the limitation. Computing cost function is a mathematical technique to measure the performance of architecture and to determine the error between actual and predicted values. The gradient descent is optimization technique to determine the optimized values of the parameters to increase the performance and minimize errors in the cost function. Although gradient descent optimization is a natural selection for optimizing the parameters, it has been identified with many limitations on their ability on non-convex complex function and on finding the global minima. Hence, Stochastic Gradient Descent (SGD) and Mini-batch Gradient Descents are developed as the optimization technique to identify the potential solution to find the global minima. Batch gradient descent optimization technique only allows the update of the parameter after the complete computation of gradient value for the entire training samples set. The stochastic gradient descent (SGD) and mini-batch gradient descent optimization allow updating the parameters on each iteration where the updates in SGD carried out on each samples iteration and mini-batch samples iterations. Mini-batch gradient descent is better than SGD on the deep network with highly redundant input data where less computation cost is required for updating the parameters and computation of the gradient can be implemented in GPU as well as in computer cluster. However, the learning rate is fixed in both SGD and mini-batch gradient descent which needs a manual tuning for steady convergence and also to reduce the computational time. Hence, several techniques are developed to accelerate the convergence such as momentum, Nesterov accelerated gradient, Adagrad, AdaDetla, RMS prop, Adam, AdaMax and Nadam.

    Regularization technique is used to avoid the overfitting of the network has more parameter than the input data and for the network learned with noisy inputs. Regularization encourages the generalization of the algorithm by avoiding the coefficient to fit so perfectly with the training data samples. To prevent the overfitting, increasing the training sample is an attractive solution. Also, data argumentation, $L_1$ and $L_2$ regularization, Dropout, Drop connect and Early stopping can be used. In ConvNets, increasing the input data, argumentation, early stopping, dropout and its variant are highly implemented for fine-tuning of the network. Model accuracy has been dramatically influenced by the selection of a set of hyperparameters, in some cases changing it from 1 to 95 [13] [14]. Over the past few years, a variety of algorithms are developed for optimizing the hyperparameters values such as grid search, random search [15], model-based approaches using random forests [16] and sequential model-based optimization [17]. The widely used optimization algorithm is grid search method. In this grid search method, the model is trained over a range of the set of hyperparameter, and the values give the best performance on the network on the cross-validation data is selected. Also, this method involves training the network with manually selected sub-set of hyperparameters and explore the defined hyperparameter space which is considered to be ineffective and extremely time-consuming. This method is severely influenced by the curse of dimensionality as the number of hyperparameters increase. The random search method is the effective alternative of grid search to explore the randomly sampled hyperparameters in the hyperparameter space. Its also time-consuming when involves a large number of hyperparameters in the search space. One of the most powerful strategies to optimize the hyperparameter is Sequential model-based optimization. In this technique involves constructing a probabilistic model to the data to determine the most promising point to evaluate. Hyperparameter optimization requires optimizing an unknown black box function. In such computationally expensive situation , Bayesian optimization technique is an efficient and powerful heuristic to optimize the function [18] [19]. Sequential model-based Bayesian Optimization is studied by many researchers and is well established global optimization strategy for unknown noisy function [20]. Bayesian optimization constructs a probabilistic surrogate model to define the distribution over the unknown black box function, and a proxy optimization is performed to seek the next location to evaluate where the posterior distribution is developed based on conditioning on the previous evaluations. Acquisition function is applied to the posterior mean and variance to express the trade-off between exploration and exploitation. Most commonly, Gaussian Process is used in the construction of the distribution over objective function because of their flexibility, well-calibrated

uncertainty, and analytic properties [21]. Bayesian Optimization is shown the better performance than grid search and random search and outperformed the state of art performance on several machine learning task [22] [23]. Swersky et al. investigation on Mutiltak Bayesian Optimization showed that the knowledge transformation between correlated tasks where they tried to determine the best configuration for the dataset by evaluating manually chosen sub-sets [24]. Nickson et at, estimated configuration performance of larger dataset by the evaluation based on several training on small random manually chosen subsets [25]. Bergstra et al. showed that complex vision architecture such as face matching verification & identification and object recognition is achieved the state of the art performance with 238 hyperparameters [26]. Also, many studies suggested that automated hyperparameter optimization is significantly improved the model performance for wide range of problems [27] [28] [29].

The complex architecture requires a large amount of data to train that leads the learning is a very computational expensive process and tends to deal with a large number of hyperparameter. Though the GPU implementation can have the capabilities of reducing the learning time, optimizing the hyperparameters are still an expensive and crucial step. This article is intended to develop an efficient and general purpose ConvNet architecture for multi-class image categorical classification application with Bayesian Optimization technique to determine the optimum value of hyperparameters. In developing the architecture, large pool of grey scale image taken as training and validation datasets. Data augmentation, $L_1$ & $L_2$ regularization, dropout and early stopping are introduced to reduce the overfitting of the architecture. Hyperparameters such as number & dimension of the convolutional layer, learning rate, the number of hidden layers, number of hidden units, $L_1$ & $L_2$ regularization, learning rate are determined by Sequential model-based Bayesian Optimization. ReLu non-linear activation function is implemented after the pooling layers. Max pooling operation is carried out to downsampling the data points in pooling layers. Cross entropy loss function is used to measure the performance of the architecture and softmax is used in the classification layer. Mini-batch gradient descent with Adam optimizer algorithm is used for backpropagation. Developed architecture is validated with confusion matrix. This article developed as a continuation of our previous study [30] [31] [32]

## 2 Architecture

Set of parallel features maps are developed by sliding different kernels over the input images and stacked together in a layer which is called as Convolutional layer. Using smaller dimension as compares with original image helps the parameters sharing between the feature maps. In the case of overlapping of the kernel with images, zero padding is used to adjust the dimension of the input images. Zero padding also introduce to control the dimensions of the convolutional layer. Activation function decides which neuron should be fired. The weighted sum of input values is passed through the activation layers. The neuron that receives the input with higher values has the higher probability of being fired. Different types of activation function are developed for the past few years that includes linear, Tanh, Sigmoid, ReLu and softmax activation functions. In practice, it is highly recommended that selection of activation function should be based deep learning framework and the field of application. Downsampling of the data is carried out in pooling layers. It reduces the data point and overfitting of the algorithm. Also, pooling layer reduces the noises in the data and smoothing the data. Usually pooling layer is implemented after the convolution process and non-linear transformations. Data points derived from the pooling layers are stretched into single column vectors and fed into the classical deep neural networks. The architecture of a typical ConvNet is given in the Figure 2.1. The cost function, also known as loss function is used to measuring the performance of the architecture by utilizing the actual $y_i$ and predicted values $\hat{y}_i$. Mean Squared Error, Mean Squared Logarithmic Error, $l_1$ $l_2$ norm, Mean Absolute percentage Error and Cross-Entropy are commonly available cost function. In practice, cross entropy is widely used in ConvNet architecture. The input image patch $z^O$ is convoluted by $D_I^l$ number of maps with size of $C_I^1 \times C_I^l$ and is produced $D_O^l$ number of output maps with the size of $C_O^l \times C_O^l$ where $z^{l-1}$ and $z^l$ are represent the input and output of the layer $l$ and $z^L$ is represent the output of the last layer $L$. The outputs of the layer $l-1$ are fed into the layer $l$ as the inputs. Hence, $j^{th}$ output feature maps of the layer $l$ is denoted by $z_j^l$ and is given as,

$$z_j^l = \sigma(\sum_i z_i^{l-1} * w_{ij}^l + b_j^l 1_{C_o^l}) \qquad \text{(Eq. 2.1)}$$

where, $0 \leq i < D_I^{(l-1)}, 0 \leq i < D_O^{l-1}$. The convolution process is indicated by the symbol $*$ and the non-linear transformation is represented by $\sigma$. The bias values $b_j^l$ are multiplied with every element of the matrix $1_{C_o^l}$ of size of $C_O^l \times C_O^l$ before added to weighted input values $\sum_i d_i^{l-1} * w_{ij}^l$. If the down-sampling is implmented in pooling layer by mean pooling, then the $(x, y)$ element ouput of the feature map $j$ of layer $l$ is expressed as,

$$z_j^l(x, y) = \frac{\sum_{m=0}^{s-1} \sum_{n=0}^{s-1} z_j^{l-1}(s \times x + m, s \times y + n)}{s^2} \qquad \text{(Eq. 2.2)}$$

where $\leq 0x$, $y < C_I^l$ and $s$ is downsampling factor. In the pooling layer. There are no weights added, and the only parameter is biased. Final layer $L$ is connected to fully connected dense layer where the vectorization and concatenation of the data point of layer $L-1$ are carried out. The output of the last layers is given by, if number of output maps $D_O^L$ for this concatenated layer is $D_O^{L-1} \times (C_O^{L-1})^2$,

$$z_j^L(0, 0) = z_i^{L-1}(x, y) \qquad \text{(Eq. 2.3)}$$

where,

$$j = i \times (C_O^{L-1})^2 + (y-1) \times C_O^{L-1} + x \qquad \text{(Eq. 2.4)}$$

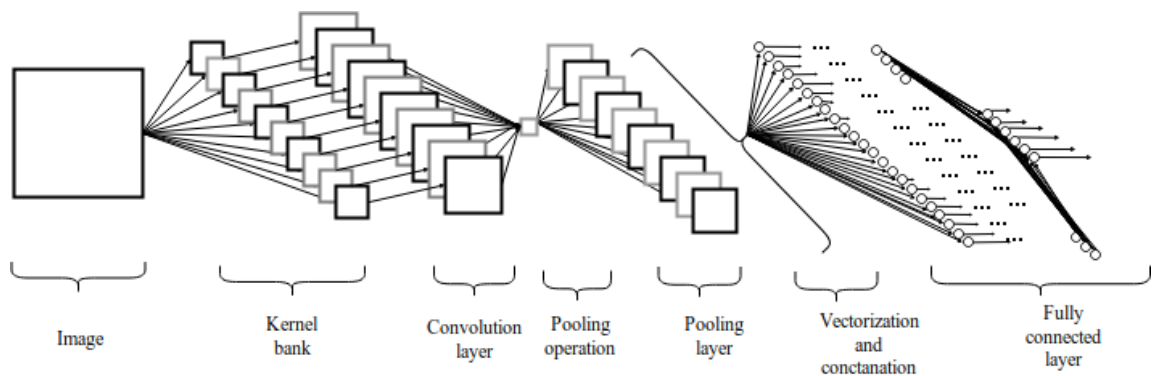

Figure 2.1: Architecture Convolution Neural Network

The probability of $m^{th}$ training samples in the class $n \in \{1, 2, \ldots K\}$ for given $K$ number of classsses and parameter matrix $\theta = (\theta_1^T, \theta_2^T \ldots \theta_K^T)$ with the size $K \times D_O^L$ is,

$$p(n|z^{L^{(m)}};\theta) = \frac{e^{\theta_n^T d^{L^m}}}{\sum_{c=1}^{K} e^{\theta_n^T d^{L^m}}} \tag{Eq. 2.5}$$

In practice, softmax classifier is widely used in multi-categorical classification problems. The parameters are identified by using maximum likelihood approach. The cost function L for the $m^{th}$ sample belongs to class $n$ is given as,

$$\mathrm{L}^m = p(n|z^{L^m};\theta) \tag{Eq. 2.6}$$

## 2.1 Loss function

Loss function maps an event of one or more variable onto a real number associated with some cost. Loss function is used to measure the performance of the model and inconsistency between actual $y_i$ and predicted value $\hat{y}_i^{L+1}$. Performance of model increses with the decrease value of loss function.

If the output vector of all possible output is $y_i = \{0, 1\}$ and an event $x$ with set of input vector variable $x = (x_i, x_2 \ldots x_t)$, then the mapping of $x$ to $y_i$ is given by,

$$\mathrm{L}(\hat{y}_i^{L+1}, y_i) = \frac{1}{t} \sum_{i=1}^{i=t} (y_i, (\sigma(x), w, b)) \tag{Eq. 2.7}$$

where $\mathrm{L}(\hat{y}_i^{L+1}, y_i)$ is loss function
Many types of loss functions are developed for various applications and some are given below.

### 2.1.1 Cross Entrophy

The most commonly used loss function is Cross Entropy loss function and is expained below. If the probablity of output $y_i$ is in the traning set label $y_i^{\hat{L}+1}$ is, $P(y_i|a^{l-1}) = i_t^{\hat{L}+1} = 1$ and the the probablity of output $y_i$ is not in the traning set label $y_i^{\hat{L}+1}$ is, $P(y_i|z^{l-1}) = y_i^{\hat{L}+1} = 0$ [33]. The expected label is $y$, than Hence,

$$P(y_i|z^{l-1}) = \hat{y}_i^{L+1} y_i (1 - \hat{y}_i^{L+1})^{(1-y_i)} \tag{Eq. 2.8}$$

$$\log P(y_i|z^{l-1}) = log((\hat{y}_i^{L+1})^{(y_t)}(1 - \hat{y}_i^{L+1})^{(1-y_i)}) \tag{Eq. 2.9}$$

$$= (y_i)\log(\hat{y}_i^{L+1}) + (1 - y_i)\log(1 - \hat{y}_t^{L+1}) \tag{Eq. 2.10}$$

To minimize the cost function,

$$\log P(y_i|z^{l-1}) = -log((\hat{y}_i^{L+1})^{(y_i)}(1 - \hat{y}_i^{L+1})^{(1-y_i)}) \tag{Eq. 2.11}$$

In case of $i$ training samples, the cost function is,

$$\mathrm{L}(\hat{y}_i^{L+1}, y_i) = -\frac{1}{t} \sum_{1}^{i=t} ((y_i)\log(\hat{y}_i^{L+1}) + (1 - y_i)\log(1 - \hat{y}_i^{L+1})) \tag{Eq. 2.12}$$

$$\tag{Eq. 2.13}$$

If the vector $t$ stores the actual true label is given as $y_i$ and predicted values $z_j^L$ is given as $\hat{y}$ for simpilicty, then the cost function can be written as,

$$L(\hat{y}_i, y_i) = \frac{1}{t} \sum_{i=1}^{i=i} f(y_i, \sigma(w_{ij} z_i + b_i)) \quad \text{(Eq. 2.14)}$$

In standard form, the paramters $\theta_i : \{w_i, b_i\}$ that minimize the distance between predicted $\hat{y}_i$ and actual $y_i$ values can be expressed as,

$$L(\theta_i) = \frac{1}{i} \sum_{i=1}^{i=i} f(y_i, \hat{y}_i) \quad \text{(Eq. 2.15)}$$

## 2.2 Mini batch Gradient descent

Gradient descent, also known as the method of steepest descent, is an iterative optimization algorithm to find the minimum of a complex function by computing successive negative derivatives points of the function through iteration process. The derivative is the rate of change of slope of a function which is typically represented the changes of a function at any given point, provided a continuous and differentiable function. Batch gradient descent also is known as vanilla gradient descent; computation is carried out to determining the gradient of the function concerning parameters. If predicted value of the out put $\hat{y}_i = f(z^L, \theta_i), \theta \in \mathbb{R}^n$ of the algorithm is obtained by passing the input value $z \in \mathbb{R}^n$ through the intermediate layers and hidden layers. From the Equations 2.6, 2.7 and 2.8, the cost function associated with that layer is given by,

$$\theta_i = \theta_i - \alpha \frac{\partial}{\partial \theta_i} L(\theta_i : (y_i, \hat{y}_i)) \quad \text{(Eq. 2.16)}$$

Where the $\alpha$ is the learning rate. The updates of the parameter are performed for all values of $i = (0, 1, 2 \ldots i)$ in the direction of steepest descent of the cost function L [34]. In batch gradient descent, computation of all gradient points for the whole training sample sets has to be carried out to perform a single update which is considered to be inefficient in large datasets. Though the gradient value is zero at the optimum for convex function, batch gradient descent susceptible to fall into local minima for non-convex functions. Mini-batch gradient descent involves that splitting the training samples into multiple mini-batch contains multiple samples instead of a single sample at every iteration. These mini batches are employed in computing the error and updating the parameter. Sum or average of the gradients for the mini batches reduces the variance as compares with stochastic optimization which leads to more stable convergence. Mini-batch gradient descent is commonly used in deep learning models. If the batch sizes are considered to be $i : i + n$, than the mini batch gradient descent is expressed as,

$$\theta_i = \theta_i - \alpha \frac{\partial}{\partial \theta_i} (\theta : y^{i:i+n}, \hat{y}^{i:i+n}) \quad \text{(Eq. 2.17)}$$

## 2.3 Adam

Adam is known as Adaptive Moment Estimation is another technique to compute the adaptive learning rates for the parameters. Similar to adadetla and RMS prop, Adam optimizers stores the exponentially decaying average of the previous squared gradient $v_i$ but also similar to momentum, it keeps the exponentially decaying average of previous historical gradients $m_i$,

$$m_{i+1} = \beta_1 m_i + (1 - \beta_1)g_i \qquad (Eq.\ 2.18)$$

$$v_{i+1} = \beta_2 v_i + (1 - \beta_2)g_i \qquad (Eq.\ 2.19)$$

Where $v_i$ and $m_i$ are the estimations of the second and first moment of the gradient respectively. During the initialization stages, first and second momentum are biased to zero. Hence, the bias-corrected first and second moments are given by,

$$\hat{m}_i = \frac{m_i}{1 - \beta_1^i} \qquad (Eq.\ 2.20)$$

$$\hat{v}_i = \frac{v_i}{1 - \beta_2^i} \qquad (Eq.\ 2.21)$$

From the above Equations, the adam update is given as,

$$\theta_{i+1} = \theta_i - \frac{\alpha}{\sqrt{\hat{v}_i} + \epsilon}\hat{m}_i \qquad (Eq.\ 2.22)$$

In practice, the values of $\beta_1$, $\beta_2$ and $\epsilon$ are set to be 0.9, 0.999 and $10^{-8}$.

# 3 Data preparation and Experimentaion

Development and implementation of the architecture are carried out by using Karas, an open source python library. Traning of the architecture is performed by using a CPU machine with an i7-950 core(3.33 GHz), 8 GB DDR3. A lot of high-resolution colour images are collected from various sources such that it can contain four categorical images. These images are downsampled to a fixed window size of $128 \times 128$ since the images have variation in the image quality and size. Images are converted into grey scale image for the computational simplicity, and data augmentation procedure is implemented. Total images are split into training and test data set with the ratio of 7:3 resulting with 3200 training set images and 800 test set images. The values given in the Table 1

| Category | Label | Training set | validation set |
|---|---|---|---|
| Human | 0 | 800 | 200 |
| Horse | 1 | 800 | 200 |
| cat | 2 | 800 | 200 |
| Dog | 3 | 800 | 200 |

Table 1: Dataset prepration

# 4 Techniques for improving the architecture

Since because of the function of being a potential tool for ensuring the generalization of the algorithm, studies on regularization of the algorithm becomes the main research topic in machine learning [35] [36]. Moreover, the regularization becomes very crucial step in the deep learning model that has more parameters than the training data sets. Regularization is a technique to avoids the overfitting of the algorithm and to avoids the overfitting of coefficients to fit so perfectly as model complexity increases. Overfitting often occurs when the algorithm learns the input data along with noises. Over the past few years, verity of methods are proposed and developed for the machine learning algorithm to regularize such as data argumentation, $L_2$ regularization or weight decay, $L_1$ regularization, dropout, drop connect, stochastic pooling and early stopping [37] [38] [39] [40] [37] [7] [41].

## 4.1 Data agumentation

To increasing the performance of the algorithm as well as satisfying the requirements of a large amount of data for the deep learning model, data augmentation is an important tool to be implemented. Data augmentation is a technique to artificially increase the training set by adding transformations or perturbations of the training data without increasing the computational cost. Data augmentation techniques such as flipping the images horizontally or vertically, crop, color jittering, scaling and rotations are commonly used in the visual imagery and image classification applications. In imagenet classification, Krizhevsky et al. [7] are proposed a methodology of implementing PCA to alter the intensity of the RGB color channels on training AlexNet and also they proposed that the PCA approximately captured the notable properties of the images. Bengio et al., proved that the deep architecture benefits more from the data augmentation technique as compares with the shallow network [42]. Zhang et al., implemented the data argumentation technique along with explicit regularizers such as weight decay and dropouts [41]. Also, it has been used in many image classification problems and found proven as a successful implementation. Chaoyun et al., successfully implemented data augmentation technique to improve the performance of leaf classification and they also claimed that developed ConvNet architecture is outperformed other classification methods [43].

## 4.2 $L_1$ and $L_2$ regularization

The most commonly used regularization methods are $L_1$ and $L_2$ regularization methods. In $L_1$ regularization, regularization term added to the objective function to reduce the sum of absolute value of the parameters wherein $L_2$ regularization, the regularization term is added to reduce the sum of the squares of the parameters. It has been understood from the previous investigations, many parameter vectors in $L_1$ regularization is sparse since the many models cause the parameter to zero. Hence, It has been implemented in feature selection setting where the many features are needed to be avoided or ignored. Most commonly used regularization method in machine learning is imposing a squared $L_2$ norm constraint on the weights. It is also known as weight decay (Tikhonov regularization) since the net effect of reducing the weight by a factor, proportional to the magnitude at every iteration of the gradient descent [44]. Enforcing the sparsity with the weight decay is to artificially introduce the zeros on all lower weights in an absolute manner than referring a threshold. Even in this, the effect of sparsity is negligible. The standard regularized cost function is given by,

$$\theta = \arg\min_{\theta} \frac{1}{N} \sum_{i=1}^{N} (L(\hat{y}_i, y) + \lambda R(w)) \qquad \text{(Eq. 4.1)}$$

where the regularization term $R(w)$ is,

$$R_{L_2}(w) \triangleq ||W||_2^2 \qquad (Eq.\ 4.2)$$

Another method is penalize the absolute magintude of the weights, known as $L_1$ regularization.

$$R_{L_1}(w) \triangleq \sum_{k=1}^{Q} ||W||_1 \qquad (Eq.\ 4.3)$$

The formulation of $L_1$ regularization is not differentiable at zero, hence weights are increased by a constant factor close to zero. It is common in many neural networks to apply the first order procedures as for the weight decay formulation to solve non-convex $L_1$ regularized problems [45]. The approximate variation of $L_1$ norm is given by,

$$|W|_1 = \sum_{k=1}^{Q} \sqrt{w_k^2 + \epsilon} \qquad (Eq.\ 4.4)$$

Another regularization method is considered the mixture of $L_1$ and $L_2$ regularization is known as elastic net penalization [46].

## 4.3 Dropout

Dropout refers to temporarily dropping out the neurons along with their connection. Randomly dropping out the neurons prevent the overfit as well as provide the possible way of combining many different network architecture exponentially and efficiently. Neurons are dropped out with the probability of $1 - p$ and reduced the co-adaptability between the neurons. On these stages, the hidden unit usually implemented with a probability of 0.5 drops out neurons. Sample average of all possible $2^n$ drop out neurons are approximately computed by using the full network work with each node's output weighted by a factor of $p$, since using the larger value of $n$ is unfeasible. Drop out significantly reduce the overfitting as well as increasing the learning speed of the algorithm by avoiding the training nodes on training data. The studies and experiments on drop out in fully connected layers by reducing the test result errors from 15.60% to 14.32 % on CIFAR-10 data sets and also implementing dropout in convolution layer reduced error to 12.61%. They also achieved the same trend of improved in the performance on SVH data sets [39]. Hinton et al., implemented the drop out in fully connected layers and they provide that the convolution shared filter architecture reduce the parameters and it improves the resistance to the overfitting in convolutional layers [47]. In AlexNet dropout, regularization technique is used along with data augmentation to reduce the overfitting where they handled 50 million parameters to classify 2200 different categorical images [7].

## 4.4 Early stopping

Early stopping provides the guidance on number passes (iteration) which are needed to be carried out to minimize the cost function. Early stopping is commonly used to prevent the poor generalization of over-expressive models on training. If the number of passes is used is too small, the algorithm tends to underfit (reduce variance but encourage bias) where if the number of passes used is too high (increase variance though reduce bias), algorithm tends to overfit. Early stopping technique is used to address this problem by determining the number of passes and eliminating the manual setting of the values. The fundamental idea behind this early stopping technique is, dividing the available data into three subsets namely, training set, validation set and test set.

The training set is used for computing the weights and bias by gradient descent optimization where the validation test is used for monitoring the training process. The computed error tends to decrease as the number of passes increases on both training and validation set. However, if the algorithm begins to overfitting the data, the validation error starts to increase. Early stopping technique provides the possibilities of stopping the passes (iteration) where the deviation in validation error began and returning the weight and bias values [48]. Early stopping is also used in boosting optimization in non-convex loss function [49] and generalization of boosting algorithms [50].

## 4.5 Hyperparameter Optimization

Bayesian optimization is one of the powerful strategy used in determining the extreme of an objective function $f(x)$ on a bounded set of $\chi \in \mathbb{R}^n$. Bayesian optimization is extremely useful in determining the extreme in black box function that does not have any expression or derivatives or in non-convex function. Bayesian optimization constructs a probabilistic model for the objective function and exploits the decision about the next promising location of the function while integrating out the uncertainty. In finding the next location to evaluate, selecting the prior over the function and acquisition functions are important that can be obtained by incorporating the prior belief about the objective function and the trade-off of exploration & exploitation.

**Definition 1.** *Bayes theorem states that the posterior probability of a model $\mathcal{M}$ given observation $\mathcal{D}$ is proportional to the likelihood of $\mathcal{D}$ given $\mathcal{M}$ multiplied by the prior probability of $\mathcal{M}$.*

$$P(\mathcal{M}|\mathcal{D}) \propto P(\mathcal{D}|\mathcal{M})P(\mathcal{M}) \qquad \text{(Eq. 4.5)}$$

The prior is represent the belief about the space of possible objective function in Bayesian Optimization. If $f(x_\ell)$ is the observation of the objection function at $i^{th}$ sample $x_\ell$, for an accumulated observation $\mathcal{D}_{1:t} = \{x_{1:t}, f(x_{1:t})\}$, the prior distribution is combined with the likelihood function $P(\mathcal{D}_{1:t}|f(x_{1:t}))$, the posterior distribution can be written as,

$$P(f(x_{1:t})|\mathcal{D}_{1:t}) \propto P(\mathcal{D}_{1:t}|f(x_{1:t}))P(f(x_{1:t})) \qquad \text{(Eq. 4.6)}$$

This posterior distribution captures the updated beliefs about the objective function. Bayesian optimization utilizes the acquisition function to find the next location point $x_{t+1} \in \mathbb{R}^d$ for sampling. This automatic representation of trade-off of explorations and exploitations help to minimize the number of location points on the objective function to evaluate. This technique can be used in the objective function with multiple local extremes. Evaluation of next point is carried out by computing the maximum in the acquisition function which can be done by accounting the mean and covariance of the predictions. Then the objective is sampled at $\arg\max x$ of the functions. The GP is updated, and the process is repeated. The prior and observation are used to define the posterior distribution over the spaces of the objective functions where the informative priors are describing the attributes of the functions include smoothness and extreme even when the function itself is not known.

A Gaussian process is the generalization of Gaussian probability distribution where the probability distribution describes random variables(scalar or vector) and the properties are governed by the stochastic process. For any given choice of distinct long input vector $x_\ell = \{x_1, x_2, \ldots x_n\}, i \in \chi$ for all $i$, the output is $f(x_\ell) = \{f(x_1), f(x_2) \ldots f(x_n)\}$ and is gaussian distributed. The Guassian distribution over a vector $f(x_\ell)$ has multivariate normal distribution with mean vector $\mu = \mathbb{E}f(x_\ell)$ and covariance matrix $\gamma = Cov(f(x_\ell), f(x_\ell))$,

$$f(x_\ell) \sim \mathcal{GP}(\mu, \gamma) \qquad \text{(Eq. 4.7)}$$

If the function is distributed as Gaussian Processes with mean function $\mu : \chi \to \mathbb{R}$ and covarianace or kernel function $\gamma : \chi \times \chi \to \mathbb{R}$, for any given input $x_\ell = \{x_1, x_2, \ldots x_n\} \in \chi$, the output function $f(x_\ell) = \{f(x_1), f(x_2) \ldots f(x_n)\}$ is gaussian distrubited with mean $\mu_i = [\mu(x_1), \mu(x_2), \mu(x_3) \ldots \mu(x_n)]$ and $n \times n$ covariance or kernel matrix $\gamma(x_\ell, x'_\ell)$, We say,

$$f(x_\ell) \sim \mathcal{GP}(\mu(x), \gamma(x, x')) \tag{Eq. 4.8}$$

where, the mean function is,

$$\mu(x_\ell) = \mathbb{E}[f(x_\ell)] \tag{Eq. 4.9}$$

The covariance function is,

$$\gamma(x_\ell, x'_\ell) = \mathbb{E}[(f(x_\ell) - \mu(x_\ell))(f(x'_\ell) - \mu(x'_\ell))^T] \tag{Eq. 4.10}$$

The covariance function $\gamma(x_\ell, x_\ell)$ for the GPs is the crucial ingredients since it defines the properties nearness or similarity such as of the evaluation points. The most common choices of covariance functions $\gamma(x, x)$ are functions of $r(x, x) = x - x$, such as the automatic relevance determination (ARD) exponentiated quadratic covariance

$$\gamma_{SE}(x, x') = \theta_0 exp(-r^2) \quad r = \sum_{d=1}^{D} (x_d - x'_d)^2/\theta_d^2 \tag{Eq. 4.11}$$

or the ARD Matern 5/2 kernel advocated for hyperparameter tunning with Bayesian optimization,

$$\gamma_{52}(x, x') = \theta_0 \left(1 + \sqrt{5r^2} + \frac{5}{3}r^2\right) exp\{-5\sqrt{5r^2}\} \tag{Eq. 4.12}$$

Such covariance functions are invariant to translations along the input space and thus are stationary. Bayesian Optimization is a general method for determining the noisy expensive blackbox function. Gaussian Process is used to define a distribution over the objective function from the input space to a loss that has to be minimized. If using the Gaussian Process surrogate, then the expected improvement acquisition function is given by,

$$\Gamma(x) = \frac{f(x_{best}) - \mu(x; \{x_n, y_m\}, \theta)}{\sigma(x; \{x_n, y_n\}, \theta)} \tag{Eq. 4.13}$$

where, $f(x_{best}$ is the lowest observed value, The expected improvement criterion is defined as,

$$\alpha_{EI}(x; \{x_n, y_n\}\theta) = \sigma(x; \{x_n, y_n\}, \theta)(\gamma(x)\Phi(r(x)) + \mathcal{N}(\gamma(x); 0, 1)) \tag{Eq. 4.14}$$

where $\Phi$ is the cumulative distribution function of a standard normal, and $\mathcal{N}(; 0, 1)$ is the density of a standard normal.

## 4.6 Confusion matrix

A confusion matrix $\mathbb{C}_{ij} \in \mathbb{N}_{\leq 0}^{\mathcal{K} \times \mathcal{K}}$ contains all possible correct and wrong classifications $K \in \mathbb{N}_{\leq 2}$ where $\mathbb{C}_{ij}$ is represent number of times the data of class $i$ are identified as class $j$ and $\mathbb{C}_{ij}$ is represent number of times the data of class $i$ are identified as class $j$. Hence, the accuracy $\frac{\sum_{i=1} c_{ii}}{\sum_{i=1}^{\mathcal{K}} \sum_{j=1}^{\mathcal{K}} \mathbb{C}_{ij}}$ of the total number of samples are can determined by the sum of $\sum_{i=1}^{\mathcal{K}} \sum_{j=1}^{\mathcal{K}} \mathbb{C}_{ij}$.

The sum $r(i) = \sum_{j=1}^{\mathcal{K}} \mathbb{C}_{ij}$ of each class $i$ are worth being investigated as they show if the clsses are skewed. If the number of samples of one class dominates the data set, then the classifier can get a high accuracy by simply always prediction the most common class. If the accuracy of the classfier is close to the priory probability of the most common techniques to deal with skewed classes might help, An automatic criterion to check for this problem is

$$\mathcal{A} \geq \frac{max(\{r(i)|i=1,2,\ldots k\})}{\sum_{i=1}^{K} r(i)} + \epsilon \qquad \text{(Eq. 4.15)}$$

where $\mathcal{A}$ is accuracy. The class wise sensitivites are given by, Other values which should be checked are the class-wise sensitivities:

$$\mathcal{S}(k) = \frac{\mathbb{C}_{kk}}{r(k)} \in [0,1] \qquad \text{(Eq. 4.16)}$$

where $\mathcal{C}_{kk}$ is number of correct identified instanaces of class $k$ and $r(k)$ is total number of instances of class $k$. If $\mathcal{S}(i)$ is much lower than $\mathcal{S}(j)$, it is an indicator that more or clearner training data is necessary for $\mathcal{S}(i)$., then the class wise confusion is given by,

$$f(k_1, k_2) = \frac{\mathbb{C}_{k_1} \mathbb{C}_{k_2}}{\sum_{j=1}^{K} C_{k_1 j}} \qquad \text{(Eq. 4.17)}$$

# 5 Results and Discussions

## 5.1 Data preprocessing and hyperparameter optimization

The large pool of images is collected from various sources such as the internet, magazine covers. The images that have the promising probabilistic information about the data are carefully chosen manually from the large pool of images. Also, data augmentation techniques such as verticle & horizontal flips, random crop, stretch and shear are initiated to increase the number of images without compromising the computational power. Sampled images are split into training data and test data such that training and test data contains 0.7% and 0.3% of total amount of the images. The training set is used for training of the architecture, and the test set is used for the validation of the architecture. For computational simplicity, grey scale images are used in the network training. The input image windows are chosen to $128 \times 128$ pixel windows including the image information and part of backgrounds. However, the windows are carefully chosen to capture the rich amount of information about the data as well as reduce the noise in the input and also to reduce the number of false positive. Image normalization procedure is performed in all of training and test set before the training of network. Batch gradient descent with the addition of Adam optimizer is used for the minimizing the error. The network is trained for 60 epochs on training and each epoch, the training set images are randomly shuffled. In practice, it is being understood that the batch size 32 is found to effective. Hence the batch size of learning is fixed to be 32. Along with the batch size hyperparameter, other hyperparameters such as types of activation layers and cost function are manually chosen. ReLU activation function is used in the intermediate layers and softmax is introduced in the classification layer of the architecture. The categorical cross-entropy loss function is used for measuring error between actual and predicted values. However, the hyperparameters such as the number of convolutional layers $ dense layers, size of the convolutional kernel, the dimension of the dense layers, dropout, weight regularization,

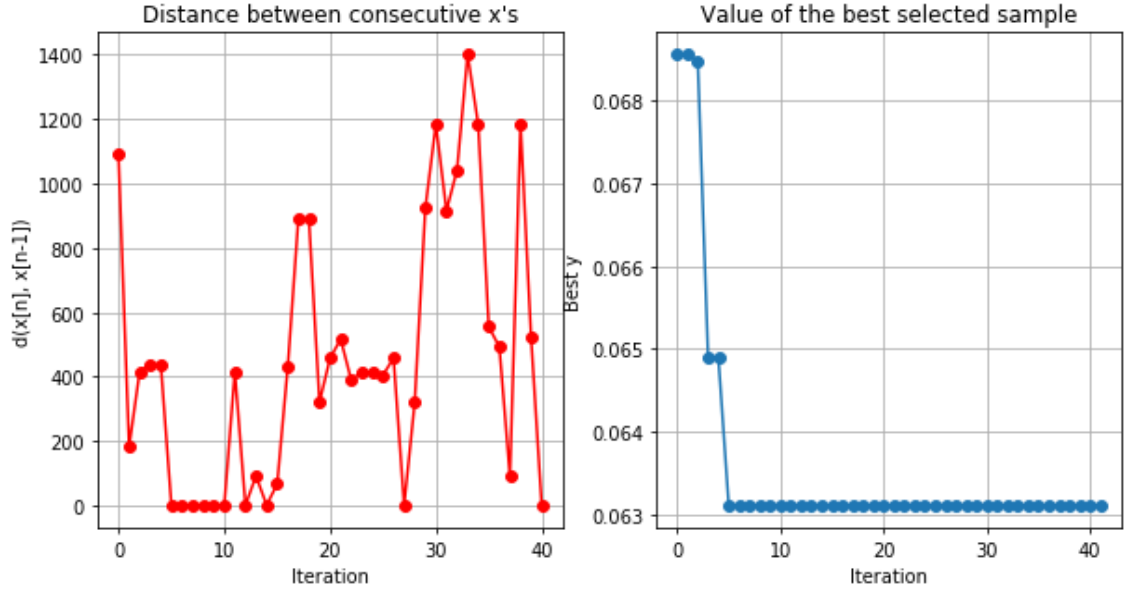

Figure 5.1: Architecture Convolution Neural Network

learning rate are determined from Bayesian Optimization algorithm. Activation layer and pooling layers are implemented after every convolutional layer. The Bayesian optimization algorithm is effectively introduced to optimize a large number of hyperparameters from the search space. For the effective implementation of the Optimization algorithm, it is common to perform $10\mathbb{N}$ trials where $\mathbb{N}$ is the number of hyperparameters. In that connection, 1000 trials are performed in tuning the hyperparameters. In the implementation of Bayesian Optimization, number of convolutional layers, pooling layers & number of dense layers are varied from 1 to 6 with the interval of 1, number of convolutional kernels is varied from 32 to 526 with the interval of 32, dense layer units are varied from 128 to 1004 with the interval of 128 and continuous type hyperparameters such as dropout, $L_1$, $L_2$ regularization and learning rate are varied from 0 to 0.2, 0 to 0.2, 0 to 0.2 and 0 to 0.2 respectively. No additional hyperparameter are considered and are taken the default value of Keras deep learning framework. From the range of values, optimum values are determined with respect to the cost function. The resultant values are given in the Table 3. The successive search points and the best-fitted values on the Bayesian Optimization are given in the Fig.5.1.

The final ConvNet architecture Fig. 2.1 is developed with four convolutional layers followed by activation layers & pooling layers, four dense layers. The dimension of the architecture is determined by the Bayesian Optimization, resulting in totally 11 layers except for the output classification layers. The size of convolutional kernels is found to be $3 \times 3$ on every convolutional layer wherein the pooling layers, fixed pixels window size of $2 \times 2$ is selected. The input image window is selected as $128 \times 128$ and is fed into the architecture for training. The output of the architecture is developed to classifying the four categorical images from the datasets. The developed architecture has approximately 8.2 million trainable parameters and has taken four days for training. The elementary features of the images are acquired in the intermediate layers, and combined complex features are found in the final convolutional layer. Since the ReLU activation function is helped to eliminate the unwanted features and the noises, only the natural information about the images is likely to be transferred through the layers. From the layer 1 to 7 contains series of successive convolutional layers and activation layer followed by pooling layers. All of the convolutional layers are developed by employing $3 \times 3$ filters, known as convolutional kernels and connected to ReLU non-linear activation functions. The output of the convolutional layers are used to form activation layers and are subsequently subsampled by using max pooling operation in pooling layers. The last four layers are fully connected dense layer which uses the extracted features from the previous layers to classifying the images. Each layer receives information from the previous layers as shown in the Fig 5.2.

| Hyperparameter | Hyperparameter Space | Type | Best fit |
|---|---|---|---|
| Input shape | [1,128,128] | Fixed | - |
| Number of Convolutional Layer | [1,2,3,4,5,6] | Discrete | [3] |
| Number of Convolutional kernel | [32,64,128,526] | Discrete | [32],[32],[64],[64] |
| Size of Convolutional kernel | (3,3) | Fixed | - |
| Activation unit in convolutional layer | ReLU | Fixed | - |
| Number of Pooling Layer | [1,2,3,4,5,6] | Discrete | [3] |
| Size of Pooling Layer | (2,2) | Fixed | - |
| Dropout | [0-0.2] | Continuous | [0] |
| Number of dense hidden layers | [1,2,3,4,5,6] | Discrete | [8] |
| Hidden layer size | [128,256,502,1004 | Discrete | [128] |
| Dropout | [0-0.2] | Continuous | [0.1] |
| Activation unit in dense layer | ReLU | Fixed | - |
| Weight regularization on dense layer, L1 | [0-0.2] | Continuous | [0.2] |
| Weight regularization on dense layer, L2 | [0-0.2] | Continuous | [0.2] |
| Learning rate | [0.01-0.00001] | Continuous | [0.00001] |
| Loss function | cross-entrophy | Fixed | - |
| Activation unit in classification layer | Softmax | Fixed | - |

Table 2: Hyperparameter search space

| Layers \ Attributes | C-size | C-Stride | C-pad | Act | P-size | P-pad | Params | O/P shape |
|---|---|---|---|---|---|---|---|---|
| conv2d | (3 × 3) | 0 | 0 | - | - | - | 320 | (32, 128, 128) |
| activation | - | - | - | ReLU | - | - | 0 | (32, 128, 128) |
| conv2d | (3 × 3) | 0 | 0 | - | - | - | 9248 | (32, 126, 126) |
| activation | - | - | - | ReLU | - | - | 0 | (32, 126, 126) |
| maxpooling2d | - | - | - | - | (2 × 2) | 0 |  | (32, 63, 63) |
| conv2d | (3 × 3) | 0 | 0 | - | - | - | 18496 | (64, 61, 61) |
| activation | - | - | - | ReLU | - | - | 0 | (64, 61, 61) |
| maxpooling2d | - | - | - | - | (2 × 2) | 0 | 0 | (64, 30, 30) |
| conv2d | (3 × 3) | 0 | 0 | - | - | - | 36928 | (64, 28, 28) |
| activation | - | - | - | ReLU | - | - | 0 | (64, 28, 28) |
| maxpooling2d | - | - | - | - | (2 × 2) | 0 | 0 | ( 64, 14, 14) |
| dropout | - | - | 0 | - | - | 0 | 0 | (64, 14, 14) |
| flatten | - | - | 0 | - | - | 0 | 0 | (12544) |
| dense | - | - | 0 | - | - | 0 | 802880 | (64) |
| activation | - | - | - | ReLU | - | - | 0 | (64) |
| dropout | - | - | 0 | - | - | 0 | 0 | (64) |
| dense | - | - | 0 | - | - | 0 | 260 | (4) |
| activation | - | - | - | Softmax | - | - | 0 | (4) |

Table 3: Learning parameters

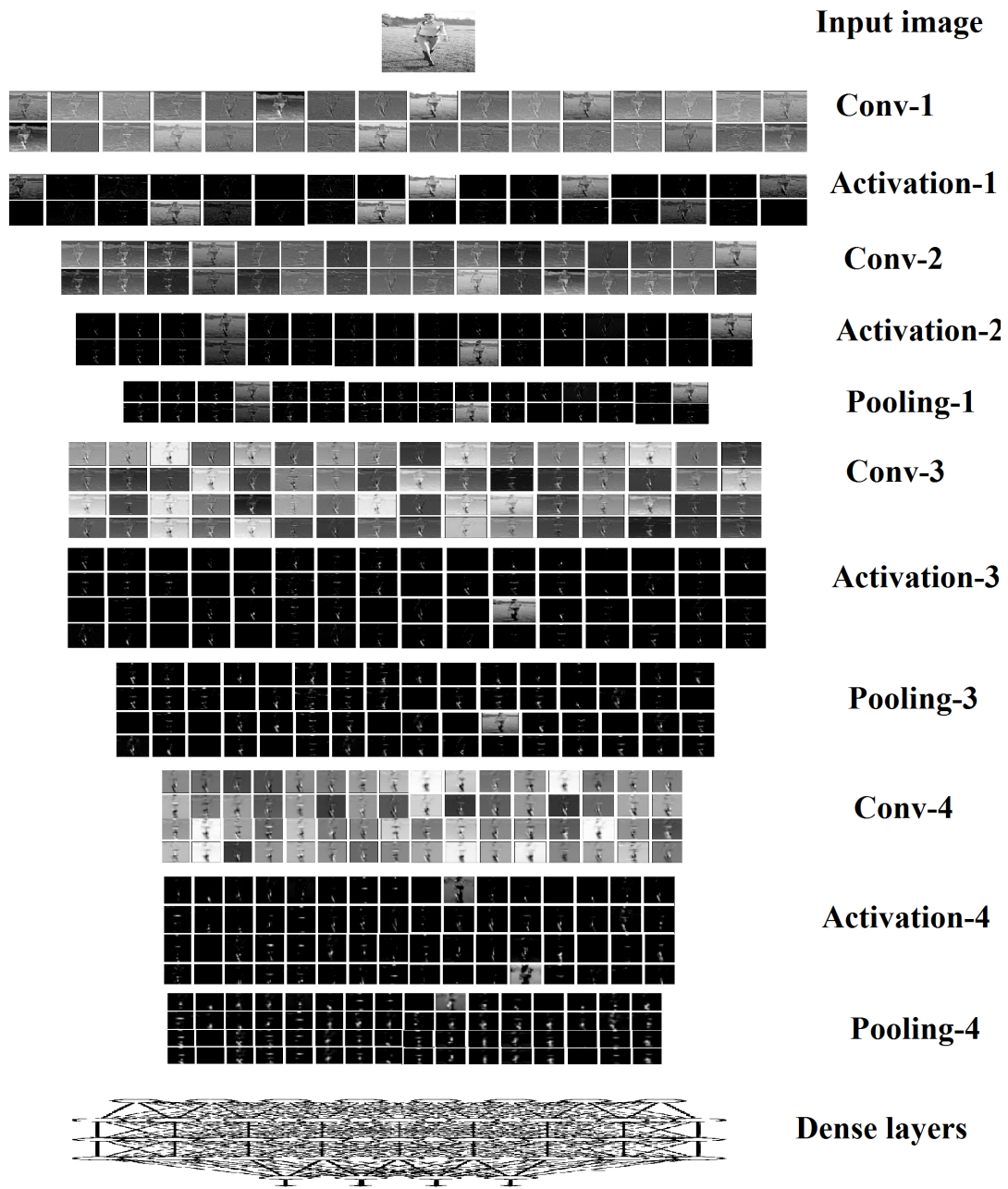

Figure 5.2: Visualization of the architecture

## 5.2 Training of the architecture

The convolutional layer-1 (Conv-1) is composed for 32 convolutional kernels with the dimension of ($32 \times 128 \times 128$). Each unit in the convolutional kernels is connected to input image neighbourhood $3 \times 3$ units, resulting in 320 trainable parameters. A weighted sum of feature maps inputs with an addition of trainable bias is computed and passed through sigmoid non-linear activation functions. Then the subsampling of the convoluted images is carried out in the pooling layer-1. The output of the pooling layer is fed into the subsequent convolutional layers 2 to convolutional layer-4. The convolutional layer-2 (Conv-2) is contained 32 convolutional kernels and formed with the dimension of $126 \times 126$. Each unit computed the weighted sum and added bias resulting with 9248 trainable parameters. Subsequently, passed through non-linear activation functions. The feature maps are subsampled of total units in the pooling layers-2. The convolutional layer-3 (conv-3) and the convolutional layer-4 (conv-4) has 64 trainable kernels and are connected to $3 \times 3$ subset units of previous layers. The weighted sum of the features maps with added bias values passed through non-linear activation function and pooling layers 3 and pooling layer four respectively. Hence, the convolutional layer-3 and convolutional layer 4 has 18496 and 36928 trainable parameters. Therefore the feature extracting layers has four convolutional layers with four activation layers followed by four activation and subsampling pooling layers where the steady features of low-dimensionality are extracted and used for classification by the fully connected dense layer. These parameters are concatenated and vectorized into single column vectors and are fed into classical Multi-Layer Perceptron, known as fully connected dense layers. From the Bayesian Optimization, it is being found, four layers of dense layers are formed with 128 units in each layer. Each layers units are dropout with 0.2 % of total units and added with $L_2$ and $L-1$ regularization. Fully connected dense layers act as a classifier, and the previous convolutional layers act feature extractors. In the fully connected dense layer-1, each of the units 128 is connected with vectorized parameters of the pooling layer-4. Also, the single unit in one dense layer is fully connected with all of the units in the previous dense layers. In these layers, the dot product is performed with weight and input values and added with bias values. The weighted sum of the input is passed through the Non-linear activation function (ReLU). Dropout is implemented in the dense layers, and the values are determined by the Bayesian Optimization algorithm. The final output of the dense layers is passed through softmax activation where categorical cross entropy is implemented to measure the error between the actual and predicted values. Hence, the fully connected dense layers have 802880 trainable parameters.nLearning of the network is carried out by the computation of Batch-gradient descent, a modified version of stochastic gradient descent algorithm. The Adam optimization algorithm is implemented to accelerate the convergence of the gradient. The performance such as the loss and the accuracy with maximum training epoch of 60 of the network is shown in the Fig.5.3.

The red curve represents the performance of the architecture on the training set, and the blue curve represents the performance of the architecture on the validation set. Early stopping algorithm is implemented to avoid the overfitting. Early stopping tends to stop the training procedure if the performance of the test data is not improved after the fixed number of iteration. It avoids overfitting by attempting to automatically select the inflection point where performance on the test dataset starts to decrease while performance on the training dataset continues to improve as the model starts to overfit. The overall accuracies of 93% and 92% for training and validation test images are acquired on the training of the architecture. Table ?? represent the confusion matrix on the test images for all of the categories. Human and horse have the highest recognition rates where the cats and dogs have the lowest recognition rates because the first pair has the distinct from the rest of the categories and second pair have the hardest classification. Majority of the misclassification occurs in that two categories. Table 4 is represent the classification report of the recognization of the datasets.

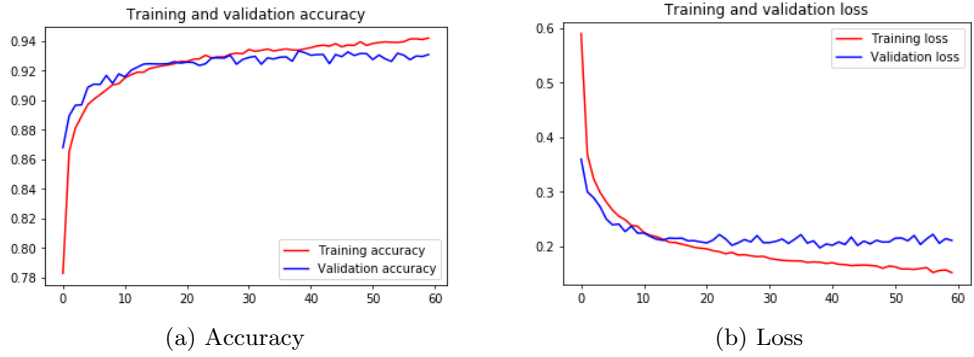

(a) Accuracy  (b) Loss

Figure 5.3: Learing curves

|  | Precision | recall | f1-score | support |
|---:|:---:|:---:|:---:|:---:|
| class-0 | 0.81 | 0.89 | 0.85 | 200 |
| class-1 | 0.99 | 0.98 | 0.99 | 200 |
| class-2 | 0.84 | 0.91 | 0.87 | 200 |
| class-3 | 0.91 | 0.93 | 0.92 | 200 |
| Avg/total | 0.92 | 0.92 | 0.92 | 800 |

Table 4: Classification report

|  | Class-0 | Class-1 | Class-2 | Class-3 |
|---:|:---:|:---:|:---:|:---:|
| **Class-0** | 188 | 5 | 3 | 4 |
| **Class-1** | 1 | 197 | 0 | 2 |
| **Class-2** | 3 | 5 | 183 | 8 |
| **Class-3** | 5 | 6 | 11 | 178 |

Table 5: Confusion matrix

# 6   Conclusion

An efficient Convolutional Neural Network for image classification is developed and presented in this article. Development of architecture and hyperparameters optimization technique are explained briefly. On developing the architecture, numerous techniques such as data augmentation, regularization, dropout, early stopping are implemented to avoid the overfitting and poor generalization. A modified version of stochastic gradient descent is used in the backpropagation of error. Softmax-Cross entropy is implemented in measuring the error between the actual and ideal values. Also, Adam optimizer is used for accelerating the convergence. Developed architecture with 11 layers has achieved the error of 92%. Results show that the developed architecture is capable of classifying the images with high accuracy.